\documentclass[times, review, 10pt]{elsarticle}

\usepackage{amsmath, bm, amssymb, amsthm}

\newtheorem{hypothesis}{Hypothesis}

\usepackage[caption=false,font=footnotesize]{subfig}
\usepackage{float}
\usepackage{threeparttable} 
\usepackage{tabularray}
\usepackage{csquotes}
\usepackage[normalem]{ulem}
\usepackage{textcomp}
\usepackage{lineno}
\usepackage{url}

\usepackage[ruled, linesnumbered, noend]{algorithm2e}
\usepackage{letltxmacro}        
\usepackage{xparse}
\SetAlCapNameFnt{\small}
\SetAlCapFnt{\small\normalfont}
\SetCommentSty{footnotesize}{}
\SetArgSty{textrm}

\ExplSyntaxOn
\let\originalSetKwInput\SetKwInput

\RenewDocumentCommand{\SetKwInput}{m m}
{
	\expandafter\NewDocumentCommand\csname #1\endcsname{O{#2} m}
	{
		\originalSetKwInput{@temp@kin}{##1}
		\csname @temp@kin\endcsname{##2}
	}
}
\ExplSyntaxOff

\SetKwInput{MyKwIn}{Input}
\SetKwInput{MyKwOut}{Output}

\makeatletter
\@ifpackagelater{algorithm2e}{2017/07/19}{}{
	\renewcommand{\algocf@Vline}[1]{
		\strut\par\nointerlineskip
		\algocf@push{\skiprule}
		\hbox{\kern-.4pt\vrule
			\vtop{\algocf@push{\skiptext}
				\vtop{\algocf@addskiptotal #1}\Hlne}}\vskip\skiphlne
		\algocf@pop{\skiprule}
		\nointerlineskip}
	\renewcommand{\algocf@Vsline}[1]{
		\strut\par\nointerlineskip
		\algocf@bblockcode%
		\algocf@push{\skiprule}
		\hbox{\kern-.4pt\vrule
			\vtop{\algocf@push{\skiptext}
				\vtop{\algocf@addskiptotal #1}}}
		\algocf@pop{\skiprule}
		\algocf@eblockcode%
	}
}
\makeatother

\LetLtxMacro\oldalgorithm\algorithm          
\LetLtxMacro\endoldalgorithm\endalgorithm    

\renewenvironment{algorithm}[1][]{%
	\oldalgorithm[#1]              
	\linespread{0.93}\selectfont   
}{%
	\endoldalgorithm               
}

\usepackage[utf8]{inputenc}     
\usepackage[T1]{fontenc}        
\usepackage{listings}           

\usepackage{courier}                

\usepackage{xcolor}       
\definecolor{codeblue}{rgb}{0,0,0.8}
\definecolor{codegreen}{rgb}{0,0.5,0}
\definecolor{codegray}{rgb}{0.5,0.5,0.5}

\journal{}

\begin{document}

\begin{frontmatter}

\title{Ada2MS: A Hybrid Optimization Algorithm Based on Exponential Mixing of Elementwise and Global Second-Moment Estimates}

\author[1]{Meng Zhu}
\ead{zhumeng@jxufe.edu.cn}
\author[1]{Quan Xiao}
\ead{xiaoquan@foxmail.com}
\author[2,3,4]{Weidong Min\corref{corresponding}}
\ead{minweidong@ncu.edu.cn}

\affiliation[1]{organization={School of Information Management and Mathematics, Jiangxi University of Finance and Economics},
	city={Nanchang},
	citysep={},
	postcode={330032}, 
	country={China}}
\affiliation[2]{organization={School of Mathematics and Computer Science, Nanchang University},
	city={Nanchang},
	citysep={},
	postcode={330031}, 
	country={China}}
\affiliation[3]{organization={Institute of Metaverse, Nanchang University},
	city={Nanchang},
	citysep={},
	postcode={330031}, 
	country={China}}
\affiliation[4]{organization={Jiangxi Provincial Key Laboratory of Virtual Reality},
	city={Nanchang},
	citysep={},
	postcode={330031}, 
	country={China}}

\cortext[corresponding]{Corresponding author}

\begin{abstract}
Optimization algorithms are core methods by which machine learning models iteratively minimize loss functions, update parameters, learn from data, and improve performance. Momentum SGD and AdamW represent two important optimization paradigms. AdamW produces stable updates and usually has strong robustness across training scenarios, but its generalization performance is sometimes weaker than that of momentum methods. Momentum SGD can often obtain better generalization after careful tuning, but it is more sensitive to gradient-scale variation and hyperparameter settings. To balance the strengths and weaknesses of the two paradigms, this paper proposes Ada2MS, an optimization algorithm that achieves a smooth transition between AdamW-like behavior and momentum-SGD-like behavior through continuous exponential interpolation between elementwise second-moment estimates and global second-moment estimates. On the visual tasks evaluated in this study, Ada2MS obtains competitive results under a unified optimizer-comparison protocol. The code will be released at \url{https://github.com/mengzhu0308/Ada2MS}.
\end{abstract}


\end{frontmatter}

\section{Introduction}\label{sec1}

Optimization algorithms are central to improving model convergence efficiency. Their significance can be summarized from three perspectives. First, they reduce hardware cost. Improved optimization algorithms can reduce the number of convergence steps and improve training efficiency under a fixed hardware budget. Second, they promote scientific discovery. Differentiable optimization frameworks play key roles in protein structure prediction~\cite{2021-Jumper-AlphaFold2}, materials design~\cite{2023-Merchant-scaling}, and controllable nuclear fusion simulation~\cite{2024-Kim-HFP}. Third, they reduce energy consumption and carbon emissions. Through algorithmic and system-engineering optimization, the pretraining of LLaMA 2-70B achieved total carbon emissions of 539 tCO$_2$e~\cite{2023-Touvron-Llama2}.

AdamW~\cite{2017-Loshchilov-AdamW} remains a widely used optimization algorithm in current deep learning. This is mainly because it can automatically handle differences in gradient scale and usually provides stable parameter updates across training scenarios. In particular, the root-mean-square norm of the AdamW update is bounded, and decoupled weight decay also helps control parameter magnitude. AdamW therefore has strong practical stability.

However, adaptive update steps may also have adverse effects. From the perspective of noise, adaptive steps introduce anisotropic noise, suppressing exploration in sharp regions and amplifying perturbations in flat regions, which makes the optimizer more likely to prefer sharp extrema~\cite{2017-Keskar-SwitchingAdamSGD,2022-Xie-AdaptiveInertia}. By contrast, stochastic gradient descent (SGD) with momentum~\cite{2013-Sutskever-SGDM}, when it can train stably, is often more likely to obtain flat solutions with better generalization. However, it cannot automatically handle gradient-scale variation and therefore often requires more careful hyperparameter tuning for recent architectures~\cite{2021-Dosovitskiy-ViT, 2022-Liu-ConvNeXt}.

Based on this trade-off, this paper proposes Ada2MS. Unlike discrete switching, Ada2MS performs smooth interpolation between AdamW-like behavior and momentum-SGD-like behavior through exponential interpolation between elementwise second-moment estimates and global second-moment estimates, as detailed in Algorithm~\ref{algo-Ada2MS}. The design aims to gradually introduce more SGD-oriented update characteristics in the later stage of training while maintaining update-scale stability as much as possible.

The remainder of this paper is organized as follows. Section~\ref{sec2} reviews existing research on optimization algorithms in deep learning. Section~\ref{sec3} presents the proposed Ada2MS algorithm in detail. Section~\ref{sec4} presents the experimental protocol, results, and analysis. The final section summarizes the paper and discusses future work.

\section{Related work}\label{sec2}

In 1986, Rumelhart \emph{et al.}~\cite{1986-Rumelhart-GD-SGD} first presented the backpropagation algorithm, which can compute exact gradients for multilayer networks over a full training set in a single pass. Although this algorithm is theoretically complete, the computation and storage cost of one iteration becomes restrictive for large-scale datasets. To overcome this limitation, Rumelhart \emph{et al.}~\cite{1986-Rumelhart-GD-SGD} also proposed SGD, which replaces full gradients with mini-batch gradient approximations. It randomly selects a mini-batch at each iteration to estimate the gradient and thus substantially reduces the cost of one parameter update. Nevertheless, the practical convergence behavior of SGD still has three main weaknesses. (1) Absence of momentum. The update direction depends completely on the current gradient, which causes persistent oscillations on narrow or high-curvature error surfaces and slows convergence. (2) Imbalanced step sizes. When the gradient scales of different parameters differ significantly, a unified learning rate cannot simultaneously accommodate sparse and dense features, causing some parameters to stop updating and others to fluctuate strongly. (3) Poor learning-rate robustness. A learning rate that is too large can make the objective diverge, whereas a learning rate that is too small can stall updates.

To address the first and third weaknesses of SGD, Sutskever \emph{et al.}~\cite{2013-Sutskever-SGDM} proposed an exponentially weighted momentum mechanism for SGD. This mechanism integrates historical gradient information into the current update direction, suppressing oscillations, accelerating movement across flat regions, and improving robustness to the learning rate. Gitman \emph{et al.}~\cite{2019-Gitman-UnderstandingSGDM} used stochastic differential equations to model how different momentum values affect escape from saddle points and convergence speed, providing practical ranges for momentum. Ramezani-Kebrya \emph{et al.}~\cite{2024-Ramezani-Kebrya-GeneralizationSGDM} derived stability and generalization-error bounds for SGD with momentum, theoretically showing that momentum not only accelerates training but can also reduce overfitting risk. Zhao \emph{et al.}~\cite{2024-Zhao-SNGDM} proposed a \enquote{normalization plus momentum} combination and proved that it can maintain an $\mathcal{O}(1 / T)$ convergence rate under anisotropic noise, thereby effectively reducing directional oscillation in SGD.

To address the second and third weaknesses of SGD, Duchi \emph{et al.}~\cite{2011-Duchi-AdaGrad} proposed AdaGrad, which uses the cumulative sum of historical squared gradients to realize per-parameter adaptive learning steps and significantly improves convergence efficiency in sparse-data scenarios. However, because AdaGrad uses the cumulative historical squared gradient as the denominator, the learning step decreases monotonically during training and approaches zero, causing late-stage updates to almost stop. To alleviate this bottleneck, Tieleman \emph{et al.}~\cite{2012-Tieleman-RMSProp} proposed RMSProp, which replaces the global accumulation mechanism with an exponential moving average of squared gradients and avoids vanishing learning steps.

Although momentum SGD and RMSProp respectively improve local weaknesses of SGD, their advantages are not combined. Adam, proposed by Kingma \emph{et al.}~\cite{2015-Kingma-Adam}, integrates the momentum mechanism and adaptive step-size strategy and proposes bias correction to offset the inherent bias of exponentially weighted averages, further improving learning-rate robustness. AMSGrad, proposed by Reddi \emph{et al.}~\cite{2018-Reddi-AMSGrad}, replaces Adam's sliding average of the second moment with the historical maximum of the gradients to avoid convergence to local extrema with poor generalization. AdamW, proposed by Loshchilov \emph{et al.}~\cite{2017-Loshchilov-AdamW}, decouples weight decay from the parameter update of the optimizer, making it a regularization term independent of the target update and thereby improving generalization and making hyperparameter tuning more robust. Adafactor, proposed by Shazeer \emph{et al.}~\cite{2018-Shazeer-Adafactor}, uses low-rank factorization of the second moment to achieve sublinear memory usage, effectively reducing memory requirements for training large models. Sophia, proposed by Liu \emph{et al.}~\cite{2024-Liu-Sophia}, presents lightweight diagonal Hessian estimates for large language model pretraining, achieving faster convergence and lower computational cost than Adam. Sophia uses a diagonal Hessian approximation instead of the full Hessian matrix, reducing computation but sacrificing curvature-correlation information among parameters. The Hessian is estimated every $k$ steps, which limits average computational overhead to within $5\%$ of gradient computation, but it may still cause delayed dynamic responses, and the estimation variance may be amplified in early training or small-batch scenarios, affecting stability.

Although Adam and its variants can converge faster to extrema, they are less likely than SGD to find flat extrema with favorable generalization. To address this limitation, Keskar \emph{et al.}~\cite{2017-Keskar-SwitchingAdamSGD} proposed \textbf{S}witching from \textbf{A}dam \textbf{t}o \textbf{S}GD (SWATS), a hybrid optimization algorithm whose key innovation is to adaptively switch from Adam to SGD during training, eliminating manual adjustment of the switching time and SGD learning rate. To address the abnormal variance of Adam's adaptive learning steps in the initial training stage caused by insufficient samples, Rectified Adam (RAdam), proposed by Liu \emph{et al.}~\cite{2020-Liu-RAdam}, mathematically estimates the confidence of the gradient degrees of freedom $\rho$. If $\rho > 4$, it enables a variance-rectified adaptive learning rate; if $\rho \le 4$, it falls back to a non-adaptive momentum mechanism. AdaX, proposed by Li \emph{et al.}~\cite{2020-Li-Adax}, changes Adam's exponential decay mechanism for the second moment into exponential growth, giving historical gradient information exponential long-term memory and thereby alleviating Adam's tendency in the stationary training phase to produce excessively large steps due to sensitivity to small gradients. Xie \emph{et al.}~\cite{2022-Xie-AdaptiveInertia} used a dynamical framework to show that Adam escapes saddle points faster but is less likely than SGD to find flat minima with favorable generalization, and they proposed the \textbf{A}daptive \textbf{I}nertia (AdaI) optimizer. Zhu \emph{et al.}~\cite{2025-Zhu-AdamNX} proposed AdamNX, whose core innovation is a new exponential decay rate for second-moment estimation, so that the strength of learning-step correction gradually weakens as training proceeds. The optimizer therefore degenerates into stochastic gradient descent with momentum in the stationary phase, improving late-stage training stability and potentially improving generalization.

In addition, both Adam and AdamW require first- and second-moment estimates to be stored for each parameter, leading to relatively high memory and computational cost. To overcome this limitation, Chen \emph{et al.}~\cite{2023-Chen-Lion} proposed Lion, which uses a sign-momentum mechanism to effectively reduce memory requirements and computation while matching or exceeding Adam on many tasks. However, when the batch size is below $64$, Lion performs worse than AdamW. More recently, Jordan \emph{et al.}~\cite{2024-Jordan-Muon} proposed Muon specifically for matrix parameters. Muon uses Newton--Schulz iteration to approximately orthogonalize the momentum update matrix, significantly improving training efficiency. Nevertheless, Muon still has several limitations: (1) its computation time is higher than Adam's; (2) when matrix shards are distributed across multiple devices, gradients from all shards must first be aggregated by all-reduce before the update can be computed, making independent parallel updates on each device difficult and introducing additional communication overhead; (3) when a model is extremely overtrained, Muon behaves similarly to Adam; and (4) Muon must be used in combination with Adam.

In summary, the development of deep learning optimizers has consistently focused on improving convergence efficiency, reducing computational cost, and enhancing hyperparameter robustness. Therefore, mixing Adam-like and SGD-like behavior is not a problem setting first introduced by this paper. The contribution boundary of this paper should be defined more precisely: Ada2MS does not claim to be the first to combine Adam-like and SGD-like optimization behavior. Instead, it realizes a smooth transition through continuous interpolation between elementwise second-moment normalization and global second-moment normalization. This differs from discrete switching methods such as SWATS and also from methods such as AdaBound that gradually approach SGD behavior through effective learning-rate bounds.

\section{Proposed optimization algorithm}\label{sec3}

\subsection{Derivation assumptions}\label{sec3.1}

Let
\begin{equation}
    \bm{g}_t = \nabla f(\bm{\theta}_{t-1}) + \bm{\xi}_t
    \label{eq-sgd-gd-noise}
\end{equation}
where $\nabla f(\bm{\theta}_{t-1})$ denotes the full gradient, namely the gradient of $\bm{\theta}_{t-1}$ computed using all samples $\bm{\mathcal{D}}$. $\bm{g}_t$ denotes the stochastic gradient, namely the gradient of $\bm{\theta}_{t-1}$ computed by randomly drawing a subset $\bm{\mathcal{R}} \subset \bm{\mathcal{D}}$. $\bm{\xi}_t$ denotes the noise introduced by random sampling. Therefore, $\bm{g}_t$ is only an approximate estimate of $\nabla f(\bm{\theta}_{t-1})$. In the experiments in Section~\ref{sec4}, the mini-batch gradient serves as this stochastic estimator. The purpose of this section is mainly to provide a mechanism-level stability analysis under simplified assumptions, rather than a complete convergence theorem for mini-batch nonconvex Ada2MS. Because sampling is random, $\bm{\xi}_t$ is isotropic, and the noises $\bm{\xi}_t$ and $\bm{\xi}_{t+i}$ at different times are independent. For the semi-qualitative and semi-quantitative analysis below, we make the following assumptions:
\begin{hypothesis}\label{hypothesis-grad-noise-iid}
The components of the full gradient are independent and identically distributed, and the components of the noise vector are independent and identically distributed.
\end{hypothesis}
\begin{hypothesis}\label{hypothesis-noise-norm}
$\bm{\xi}_t \sim \mathcal{N}(\mathbf{0}, \sigma^2\mathbf{1})$.
\end{hypothesis}
\begin{hypothesis}\label{hypothesis-grad-exp-var-stationarity}
The full gradient is a random variable and is stationary.
\end{hypothesis}
\begin{hypothesis}\label{hypothesis-grad-noise-independent}
The full gradient $\nabla f(\bm{\theta}_{t-1})$ and the random noise $\bm{\xi}_t$ are independent.
\end{hypothesis}
Based on Assumptions~\ref{hypothesis-grad-noise-iid} and~\ref{hypothesis-noise-norm}, Equation~\eqref{eq-sgd-gd-noise} can be written as
\begin{equation}
    \bm{g}_t = \nabla f(\bm{\theta}_{t-1}) + \sigma \bm{\xi}
    \label{eq-sgd-gd-noise-S2}
\end{equation}
where $\bm{\xi}$ denotes the standard normal distribution. Based on Assumptions~\ref{hypothesis-grad-noise-iid} and~\ref{hypothesis-grad-exp-var-stationarity}, we have
\begin{equation}
\begin{cases}
    \mathbb{E}_t[\nabla f(\bm{\theta}_{t-1})] = \mu_{\nabla f} \mathbf{1},
    \\
    \mathrm{Var}_t[\nabla f(\bm{\theta}_{t-1})] = \sigma_{\nabla f}^2 \mathbf{1}
\end{cases}
    \label{eq-grad-exp-var-stationarity}
\end{equation}
Based on Assumptions~\ref{hypothesis-grad-noise-iid} and~\ref{hypothesis-grad-noise-independent}, we have
\begin{equation}
    \mathrm{Cov}[\nabla f(\bm{\theta}_{t-1}), \bm{\xi}_t] = \mathbf{0}
    \label{eq-grad-noise-independent}
\end{equation}

\subsection{Statistical dimensionality of momentum SGD}\label{sec3.2}

For momentum SGD\textbf{W}, the update rule is
\begin{equation}
    \begin{cases}
        \bm{m}_t = \beta \bm{m}_{t - 1} + \bm{g}_t,
        \\
        \bm{\theta}_t = \bm{\theta}_{t - 1} -
        \eta_t (\bm{m}_t + \lambda \bm{\theta}_{t - 1})
    \end{cases}
    \label{eq-sgdwm}
\end{equation}
where $\beta$ denotes the momentum coefficient, $\eta_t$ denotes the learning rate, and $\lambda$ denotes the weight-decay rate. Iterating $\bm{m}_t$ gives
\begin{equation}
    \bm{m}_t = \sum_{i = 0}^{t-1} \beta^i \bm{g}_{t-i}
    \label{eq-sgdwm-mt-iter}
\end{equation}
Substituting Equation~\eqref{eq-sgd-gd-noise-S2} into Equation~\eqref{eq-sgdwm-mt-iter} gives
\begin{equation}
    \bm{m}_t =
    \sum_{i = 0}^{t-1} \beta^i (\nabla f(\bm{\theta}_{t-1-i}) + \sigma \bm{\xi})
    \label{eq-sgdwm-mt-iter-S2}
\end{equation}
The expectation of the momentum term is
\begin{equation}
    \begin{aligned}
        \mathbb{E}[\bm{m}_t] &=
        \mathbb{E}\left[
        \sum_{i = 0}^{t-1} \beta^i (\nabla f(\bm{\theta}_{t-1-i}) + \sigma \bm{\xi})
        \right]
        = \sum_{i = 0}^{t-1} \beta^i \mathbb{E}[\nabla f(\bm{\theta}_{t-1-i}) + \sigma \bm{\xi}]
        \\
        &= \sum_{i = 0}^{t-1} \beta^i \mathbb{E}[\nabla f(\bm{\theta}_{t-1-i})]
        = \frac{1 - \beta^t}{1 - \beta} \mu_{\nabla f} \mathbf{1}
    \end{aligned}
    \label{eq-sgdwm-mt-exp}
\end{equation}
The variance of the momentum term is
\begin{equation}
    \begin{aligned}
        \mathrm{Var}[\bm{m}_t] &= \sum_{i = 0}^{t-1} \beta^{2i}
        \mathrm{Var}[\nabla f(\bm{\theta}_{t-1-i}) + \sigma \bm{\xi}]
        \\
        &= \sum_{i = 0}^{t-1} \beta^{2i} (
        \mathrm{Var}[\nabla f(\bm{\theta}_{t-1-i})] +
        \mathrm{Var}[\sigma \bm{\xi}] +
        2\mathrm{Cov}[\nabla f(\bm{\theta}_{t-1-i}), \sigma \bm{\xi}]
        )
        \\
        &= \sum_{i = 0}^{t-1} \beta^{2i}
        (\sigma_{\nabla f}^2 + \sigma^2) \mathbf{1}
        = \frac{1 - \beta^{2t}}{1 - \beta^2} (\sigma_{\nabla f}^2 + \sigma^2) \mathbf{1}
    \end{aligned}
    \label{eq-sgdwm-mt-var}
\end{equation}
The root-mean-square (RMS) norm of the momentum term is
\begin{equation}
    \begin{aligned}
        \lVert \bm{m}_t \rVert_{\mathrm{RMS}} &=
        \sqrt{\frac{1}{d} \mathbb{E}[\lVert \bm{m}_t \rVert^2]}
        = \sqrt{\frac{1}{d} \mathbb{E}\left[\sum_{i=1}^{d}m_{t,i}^2\right]}
        = \sqrt{\frac{1}{d} \sum_{i=1}^{d}\mathbb{E}[m_{t,i}^2]}
        \\
        &= \sqrt{\mathbb{E}[m_{t,i}^2]}
        = \sqrt{
        \left(
        \frac{1 - \beta^t}{1 - \beta} \mu_{\nabla f}
        \right)^2 +
        \frac{1 - \beta^{2t}}{1 - \beta^2} (\sigma_{\nabla f}^2 + \sigma^2)
        }
    \end{aligned}
    \label{eq-sgdm-update-rms}
\end{equation}
Here $d$ denotes the vector dimension.

\subsection{Statistical dimensionality of AdamW}\label{sec3.3}

For Adam\textbf{W}, the update rule is
\begin{equation}
    \begin{cases}
        \bm{m}_t = \beta_1 \bm{m}_{t - 1} + (1 - \beta_1)\bm{g}_t,
        \\
        \bm{v}_t = \beta_2 \bm{v}_{t - 1} + (1 - \beta_2)\bm{g}_t^2,
        \\
        \overset{\frown}{\bm{m}}_t = \frac{\bm{m}_t}{1 - \beta_1^t},
        \quad
        \overset{\frown}{\bm{v}}_t = \frac{\bm{v}_t}{1 - \beta_2^t},
        \\
        \bm{\theta}_t = \bm{\theta}_{t - 1} -
        \eta_t
        \left(
        \frac{\overset{\frown}{\bm{m}}_t}{\epsilon + \sqrt{\overset{\frown}{\bm{v}}_t}} +
        \lambda \bm{\theta}_{t - 1}
        \right)
    \end{cases}
    \label{eq-adam}
\end{equation}
where $\beta_1$ and $\beta_2$ denote the exponential decay rates for the first- and second-moment estimates, respectively, and $\epsilon$ is a small constant used to prevent division by zero. Iterating $\bm{m}_t$ gives
\begin{equation}
    \bm{m}_t = (1 - \beta_1) \sum_{i = 0}^{t-1} \beta_1^i \bm{g}_{t-i}
    = (1 - \beta_1) \sum_{i = 0}^{t-1} \beta_1^i
    (\nabla f(\bm{\theta}_{t-1-i}) + \sigma \bm{\xi})
    \label{eq-adam-mt-iter}
\end{equation}
The expectation of the first-moment estimate in AdamW is
\begin{equation}
    \mathbb{E}[\bm{m}_t] = (1 - \beta_1) \sum_{i = 0}^{t-1} \beta_1^i
    \mathbb{E}[\nabla f(\bm{\theta}_{t-1-i})]
    = (1 - \beta_1^t) \mu_{\nabla f} \mathbf{1}
    \label{eq-adam-mt-exp}
\end{equation}
The variance of the first-moment estimate in AdamW is
\begin{equation}
    \begin{aligned}
        \mathrm{Var}[\bm{m}_t] &= (1 - \beta_1)^2 \sum_{i = 0}^{t-1} \beta_1^{2i}
        \left(
        \mathrm{Var}[\nabla f(\bm{\theta}_{t-1-i})] +
        \sigma^2 \mathbf{1}
        \right)
        \\
        &= \frac{1 - \beta_1}{1 + \beta_1}
        (1 - \beta_1^{2t})
        (\sigma_{\nabla f}^2 + \sigma^2) \mathbf{1}
    \end{aligned}
    \label{eq-adam-mt-var}
\end{equation}
Iterating $\bm{v}_t$ gives
\begin{equation}
    \begin{aligned}
        \bm{v}_t &= (1 - \beta_2) \sum_{i = 0}^{t-1} \beta_2^i \bm{g}_{t-i}^2
        = (1 - \beta_2) \sum_{i = 0}^{t-1} \beta_2^i
        (\nabla f(\bm{\theta}_{t-1-i}) + \sigma \bm{\xi})^2
        \\
        &= (1 - \beta_2) \sum_{i = 0}^{t-1} \beta_2^i
        \left(
        (\nabla f(\bm{\theta}_{t-1-i}))^2 +
        2 \nabla f(\bm{\theta}_{t-1-i}) \odot \sigma \bm{\xi} +
        \sigma^2 \bm{\xi}^2
        \right)
    \end{aligned}
    \label{eq-adam-vt-iter}
\end{equation}
The expectation of the second-moment estimate in AdamW is
\begin{equation}
    \mathbb{E}[\bm{v}_t] = (1 - \beta_2^t)
    (\mu_{\nabla f}^2 + \sigma_{\nabla f}^2 + \sigma^2) \mathbf{1}
    \label{eq-adam-vt-exp}
\end{equation}
The RMS norm of the AdamW update vector, excluding the update vector introduced by weight decay, is
\begin{equation}
    \begin{aligned}
        \left\lVert \frac{\overset{\frown}{\bm{m}}_t}
        {\epsilon + \sqrt{\overset{\frown}{\bm{v}}_t}} \right\rVert_{\mathrm{RMS}}
        &= \sqrt{\frac{1}{d} \mathbb{E}\left[\left\lVert
            \frac{\overset{\frown}{\bm{m}}_t}
            {\epsilon + \sqrt{\overset{\frown}{\bm{v}}_t}}
             \right\rVert^2\right]}
        = \sqrt{
        \frac{1}{d}
        \sum_{i=1}^{d} \mathbb{E}\left[
        \left(\frac{\overset{\frown}{m}_{t,i}}
        {\epsilon + \sqrt{\overset{\frown}{v}_{t,i}}}\right)^2
        \right]
        }
        \\
        &\approx \sqrt{
            \frac{1}{d}
            \sum_{i=1}^{d}
            \frac{\mathbb{E}[\overset{\frown}{m}_{t,i}^2]}
            {\left(\epsilon + \sqrt{\mathbb{E}[\overset{\frown}{v}_{t,i}]}\right)^2}
        }
        \\
        &=\frac{\sqrt{\frac{(1-\beta_1)(1+\beta_1^t)}{(1+\beta_1)(1-\beta_1^t)}
                (\sigma_{\nabla f}^2 + \sigma^2) +
                \mu_{\nabla f}^2}}
        {\epsilon + \sqrt{\sigma_{\nabla f}^2 + \sigma^2 + \mu_{\nabla f}^2}}
        \\
        &\approx \frac{\sqrt{\frac{(1-\beta_1)(1+\beta_1^t)}{(1+\beta_1)(1-\beta_1^t)}
                (\sigma_{\nabla f}^2 + \sigma^2) +
                \mu_{\nabla f}^2}}
        {\sqrt{\sigma_{\nabla f}^2 + \sigma^2 + \mu_{\nabla f}^2}}
    \end{aligned}
    \label{eq-adam-update-rms}
\end{equation}

\subsection{Comparing the update stability of momentum SGD and Adam from the RMS norm}\label{sec3.4}

Define the generalized signal-to-noise ratio (SNR) as
\begin{equation}
    \mathrm{SNR} = \frac{\mu_{\nabla f}^2}{\sigma_{\nabla f}^2 + \sigma^2}
    \label{eq-snr}
\end{equation}
Equation~\eqref{eq-snr} measures the strength of the full-gradient mean relative to the total fluctuation of the stochastic gradient, namely the proportion associated with the true signal direction. Substituting the SNR into Equation~\eqref{eq-sgdm-update-rms} gives
\begin{equation}
    \lVert \bm{m}_t \rVert_{\mathrm{RMS}} =
    \sigma_{\bm{g}}
    \sqrt{\left(\frac{1 - \beta^t}{1 - \beta}\right)^2 \mathrm{SNR} +
        \frac{1 - \beta^{2t}}{1 - \beta^2}}
    \label{eq-sgdm-update-rms-snr}
\end{equation}
It is easy to obtain
$\lim\limits_{\mathrm{SNR} \to \infty} = \infty$.
Therefore, the RMS norm of the momentum term has no finite upper bound, and momentum SGDW may be unstable during training. Substituting the SNR into Equation~\eqref{eq-adam-update-rms} gives
\begin{equation}
    \left\lVert \frac{\overset{\frown}{\bm{m}}_t}
    {\epsilon + \sqrt{\overset{\frown}{\bm{v}}_t}} \right\rVert_{\mathrm{RMS}} = \frac{\sqrt{\frac{(1-\beta_1)(1+\beta_1^t)}{(1+\beta_1)(1-\beta_1^t)}  + \mathrm{SNR}}}
    {\sqrt{1 + \mathrm{SNR}}}
    \label{eq-adam-update-rms-snr}
\end{equation}
It is easy to obtain that
$\left\lVert \frac{\overset{\frown}{\bm{m}}_t}
{\epsilon + \sqrt{\overset{\frown}{\bm{v}}_t}} \right\rVert_{\mathrm{RMS}}$
increases monotonically with SNR and has range $(\sqrt{\frac{1-\beta_1}{1+\beta_1}}, 1)$. Therefore, the RMS norm of the AdamW update vector has a finite upper bound and is more robust to SNR. AdamW is therefore more stable during training.

In summary, from the perspective of the RMS norm, AdamW updates are more stable and more robust to noise than momentum SGDW updates. However, if both optimizers can train stably, AdamW tends to converge to sharp points with slightly worse generalization, whereas momentum SGDW tends to converge to flat points with slightly better generalization. The proof is not expanded here; readers may refer to~\cite{2017-Keskar-SwitchingAdamSGD,2022-Xie-AdaptiveInertia}. Therefore, momentum SGDW and AdamW have different advantages in different training scenarios.

\subsection{Proposed Ada2MS optimizer}\label{sec3.5}

To address the limited robustness of momentum SGD and the tendency of AdamW to prefer sharp solutions, this paper proposes Ada2MS, which aims to flexibly control the mixing ratio between AdamW-like updates and momentum-SGD-like updates, as shown in Algorithm~\ref{algo-Ada2MS}. When $\alpha_t = 1$, Ada2MS degenerates into AdamW. When $\alpha_t = 0$, Ada2MS degenerates into momentum SGDW. When $\alpha_t \in (0, 1)$, Ada2MS continuously interpolates between the two.

As $\alpha_t$ decays from $1$ to $0$, the momentum-SGD component gradually increases, whereas the RMS norm of the pure momentum-SGD update has no upper bound. To alleviate this issue, this paper proposes a global second-moment estimate:
\begin{equation}
    n_{l,t} = \overset{\frown}{\beta}_{2,t} n_{l,t-1} +
    (1 - \overset{\frown}{\beta}_{2,t}) \lVert \bm{g}_{l,t} \rVert_2^2,\;
    \overset{\frown}{\beta}_{2,t} = \frac{\beta_2 - \beta_2^t}{1 - \beta_2^t}
    \label{eq-global-som}
\end{equation}
Therefore, the switching term is not written as $\bm{s}_{l,t} = \left(\bm{v}_{l,t}\right)^{\alpha_t}$, but is instead defined as a mixture of the elementwise second moment and the global second moment:
\begin{equation}
    \bm{s}_{l,t} =
    \left(\bm{v}_{l,t}\right)^{\alpha_t}
    \left(\frac{n_{l,t}}{\lvert\bm{g}_{l,t}\rvert}\right)^{1-\alpha_t}
    \label{eq-som-gsom}
\end{equation}
Equation~\eqref{eq-som-gsom} also helps align the effective update scale during the transition and reduces oscillations caused by abrupt switching.

We next analyze why, under the above assumptions, introducing the global second-moment estimate helps keep the RMS norm of the update bounded for all $\alpha_t \in [0, 1]$. Consider the extreme case $\alpha_t=0$. If the RMS norm of the update can be controlled in this case, the intermediate cases are usually easier to control. For simplicity, the core update steps at $\alpha_t=0$ are first written and rewritten with bias-correction terms:
\begin{equation}
    \begin{cases}
        \bm{m}_t = \beta_1 \bm{m}_{t-1} + (1 - \beta_1) \bm{g}_t,
        \\
        n_t = \beta_2 n_{t-1} + (1 - \beta_2) \lVert \bm{g}_t \rVert_2^2,
        \\
        \overset{\frown}{\bm{m}}_t = \frac{\bm{m}_t}{1 - \beta_1^t},
        \quad
        \overset{\frown}{n}_t = \frac{n_t}{1 - \beta_2^t},
        \\
        \bm{\theta}_t = \bm{\theta}_{t-1} -
        \eta_t \frac{\overset{\frown}{\bm{m}}_t}{\epsilon + \sqrt{\frac{\overset{\frown}{n}_t}{d}}}
    \end{cases}
    \label{eq-ada2ms-ms}
\end{equation}
Computing $\mathbb{E}[\lVert \bm{g}_t \rVert^2]$ gives
\begin{equation}
    \begin{aligned}
        \mathbb{E}[\lVert \bm{g}_t \rVert^2] &=
        \mathbb{E}[\lVert \nabla f(\bm{\theta}_{t-1}) \rVert^2] +
        \sigma^2 \mathbb{E}[\lVert \bm{\xi} \rVert^2] +
        2\sigma \mathbb{E}[\langle\nabla f(\bm{\theta}_{t-1}), \bm{\xi}\rangle]
        \\
        &= d(\sigma_{\nabla f}^2 + \mu_{\nabla f}^2) + d\sigma^2
    \end{aligned}
    \label{eq-l2-exp}
\end{equation}
The recurrence of the expectation of the global second-moment estimate is
\begin{equation}
    \mathbb{E}[n_t] = \beta_2 \mathbb{E}[n_{t-1}] + (1 - \beta_2) \mathbb{E}[\lVert \bm{g}_t \rVert^2]
    \label{eq-gsom-exp}
\end{equation}
Iterating Equation~\eqref{eq-gsom-exp} gives
\begin{equation}
    \mathbb{E}[n_t] = (1 - \beta_2^t) (d(\sigma_{\nabla f}^2 + \mu_{\nabla f}^2) + d\sigma^2)
    \label{eq-gsom-exp-S2}
\end{equation}
Therefore, the RMS norm of the update is
\begin{equation}
    \left\lVert \frac{\overset{\frown}{\bm{m}}_t}
    {\epsilon + \sqrt{\overset{\frown}{n}_t / d}} \right\rVert_{\mathrm{RMS}} \approx
    \frac{\sqrt{\frac{(1-\beta_1)(1+\beta_1^t)}{(1+\beta_1)(1-\beta_1^t)}
            (\sigma_{\nabla f}^2 + \sigma^2) +
            \mu_{\nabla f}^2}}
    {\sqrt{\sigma_{\nabla f}^2 + \sigma^2 + \mu_{\nabla f}^2}}
    \label{eq-ada2ms-ms-update-rms}
\end{equation}
Equation~\eqref{eq-ada2ms-ms-update-rms} has the same form as Equation~\eqref{eq-adam-update-rms}. Under the assumptions above, this indicates that the proposed global second-moment estimate helps keep the RMS norm of the Ada2MS update bounded for all $\alpha_t \in [0, 1]$. It should be emphasized that this section mainly provides a mechanism-level stability analysis, rather than a complete convergence theorem for mini-batch nonconvex Ada2MS.

Switching-exponent schedule. Compared with AdamW, Ada2MS has an additional hyperparameter $\alpha_t$, which is set according to Equation~\eqref{eq-alpha-t}:
\begin{equation}
    \alpha_t = \begin{cases}
    1, &t \le 0.6T,
    \\
    1 - \frac{t - 0.6T}{0.4T}, &t > 0.6T
    \end{cases}
    \label{eq-alpha-t}
\end{equation}
Equation~\eqref{eq-alpha-t} gives a simple empirical setting rather than a theoretically optimal schedule. The intuition is to maintain strong AdamW-like behavior in the early and middle stages of training for rapid search, and then gradually introduce stronger momentum-SGD-like behavior in the later stage for finer-grained search. A more complete experimental study should also include sensitivity analysis of the switching start point.

\begin{algorithm}[H]
    \caption{Ada2MS optimizer}
    \label{algo-Ada2MS}
    \MyKwIn[Input$_1$]{Training sample set $\bm{\mathcal{D}}$.}
    \MyKwIn[Input$_2$]{Parameters to be optimized $(\bm{\theta}_1, \bm{\theta}_2, \dots, \bm{\theta}_L)$, first moments $(\bm{m}_1, \bm{m}_2, \dots, \bm{m}_L)$, and second moments $(\bm{v}_1, \bm{v}_2, \dots, \bm{v}_L)$.}
    \MyKwIn[Input$_3$]{Learning rate $\eta_t$, weight-decay rate $\lambda$, exponential decay rates $(\beta_1, \beta_2) = (0.9, 0.99)$, $\epsilon = 1 \times 10^{-12}$, switching exponent $\alpha_t$, and total number of iterations $T$.}
    \MyKwOut{Extremal point $(\bm{\theta}_1^*, \bm{\theta}_2^*, \dots, \bm{\theta}_L^*)$.}
    \BlankLine
    \For{$t = 1$ \KwTo $T$}{
        Randomly draw a training subset: $\bm{\mathcal{R}}_t \subseteq \bm{\mathcal{D}}$\;
        \tcc{Forward propagation, where $\lvert\bm{\mathcal{R}}_t\rvert$ denotes the random batch size}
        Compute the loss: $\mathcal{L}_t = \frac{1}{\lvert\bm{\mathcal{R}}_t\rvert}\sum_{(\bm{x}_i, \bm{y}_i) \in \bm{\mathcal{R}}_t}
        E(\overset{\frown}{\bm{y}}(\bm{x}_i, \bm{\theta}_{1,t - 1}, \dots, \bm{\theta}_{L,t-1}), \bm{y}_i)$\;
        \tcc{Backward propagation}
        \For{$l = 1$ \KwTo $L$}{
            Compute the gradient: $\bm{g}_{l,t} = \nabla_{\bm{\theta}_{l,t-1}}\mathcal{L}$\;
            \tcc{$\overset{\frown}{\bm{m}}_{l,t} = \frac{\bm{m}_{l,t}}{1 - \beta_1^t}$, and $\bm{m}_{l,t} = \beta_1 \bm{m}_{l,t-1} + (1 - \beta_1) \bm{g}_{l,t}$ is equivalent to $\bm{m}_{l,t} = \overset{\frown}{\beta}_{1,t} \bm{m}_{l,t-1} + (1 - \overset{\frown}{\beta}_{1,t}) \bm{g}_{l,t}$, where $\overset{\frown}{\beta}_{1,t} = \frac{\beta_1 - \beta_1^t}{1 - \beta_1^t}$. The same holds for second-moment estimation}
            First-moment estimate: $\bm{m}_{l,t} = \overset{\frown}{\beta}_{1,t} \bm{m}_{l,t-1} + (1 - \overset{\frown}{\beta}_{1,t}) \bm{g}_{l,t}$, $\overset{\frown}{\beta}_{1,t} = \frac{\beta_1 - \beta_1^t}{1 - \beta_1^t}$\;
            Elementwise second-moment estimate: $\bm{v}_{l,t} =  \overset{\frown}{\beta}_{2,t} \bm{v}_{l,t-1} + (1 - \overset{\frown}{\beta}_{2,t})  \bm{g}_{l,t}^2$, $\overset{\frown}{\beta}_{2,t} = \frac{\beta_2 - \beta_2^t}{1 - \beta_2^t}$\;
            Global second-moment estimate: $n_{l,t} = \overset{\frown}{\beta}_{2,t} n_{l,t-1} + (1 - \overset{\frown}{\beta}_{2,t}) \lVert \bm{g}_{l,t} \rVert_2^2 $, $\overset{\frown}{\beta}_{2,t} = \frac{\beta_2 - \beta_2^t}{1 - \beta_2^t}$\;
            \tcc{$\lvert\bm{g}_{l,t}\rvert$ denotes the dimension of $\bm{g}_{l,t}$}
            Switching term:
            $\bm{s}_{l,t} =
            \left(\bm{v}_{l,t}\right)^{\alpha_t}
            \left(\frac{n_{l,t}}{\lvert\bm{g}_{l,t}\rvert}\right)^{1-\alpha_t}$\;
            Parameter update: $\bm{\theta}_{l,t} = \bm{\theta}_{l,t-1} -
            \eta_t \left(\frac{\bm{m}_{l,t}}{\epsilon + \sqrt{\bm{s}_{l,t}}} + \lambda_l \bm{\theta}_{l,t-1}\right)$, \newline
            $\lambda_l = \begin{cases}
                \lambda, &\text{if }\bm{\theta}_l\text{ is a matrix parameter}, \\
                0, &\text{if }\bm{\theta}_l\text{ is not a matrix parameter}
            \end{cases}$\;
        }
        \tcc{Save the best loss value and extremal point}
        \If{$\mathcal{L}_t < \mathcal{L}_0$}{
            $\mathcal{L}_0 \gets \mathcal{L}_t$\;
            \For{$l = 1$ \KwTo $L$}{
                $\bm{\theta}_l^* \gets \bm{\theta}_{l,t}$\tcp*{Store the extremal point}
            }
        }
    }
\end{algorithm}

\section{Experimental results and analysis}\label{sec4}

We conducted systematic benchmark comparisons between Ada2MS and several optimization algorithms to evaluate its behavior under a unified optimizer-comparison protocol. Experiments were performed on three benchmark datasets: CIFAR-100~\cite{2009-Krizhevsky-CIFAR-100} for image classification, PASCAL VOC~\cite{2011-Everingham-VOC2012} for object detection, and Semantic Boundaries~\cite{2011-Hariharan-SBD} for semantic segmentation. SwinV2~\cite{2022-Liu-SwinV2} was selected as the standard baseline for image classification, YOLOv7-tiny~\cite{2023-Wang-YOLOv7} was selected as the benchmark model for object detection, and U-Net~\cite{2015-Ronneberger-U-Net} was selected as the baseline architecture for semantic segmentation.

\subsection{Reproduction details}\label{sec4.1}

This section presents the experimental details to support reproducibility. Table~\ref{tbl-software-hardware-configuration} lists the main software and hardware configurations used in the experiments.

\begin{table}[H]
    \centering
    \caption{Main software and hardware configurations used in the experiments}
    \label{tbl-software-hardware-configuration}
    \begin{tblr}{
            width=\textwidth,
            colspec={X[l] X[l]},
            column{1}={font=\bfseries},
            hline{1,Z}={1pt},
            columns={valign=m},
            rowsep=0.5pt,
            cells={font=\footnotesize}
        }
        Torch~\cite{2019-Paszke-PyTorch} & 2.9.0 \\
        Torchvision~\cite{2019-Paszke-PyTorch} & 0.24.0 \\
        Python & 3.13.7 \\
        NVIDIA CUDA Toolkit & 13.0 \\
        RAM & 192 GB \\
        CPU & Intel(R) Core i9-14900KF \\
        GPU & NVIDIA RTX A5000 \\
    \end{tblr}
\end{table}

\begin{figure}[H]
    {\centering
        \includegraphics[width=\textwidth]{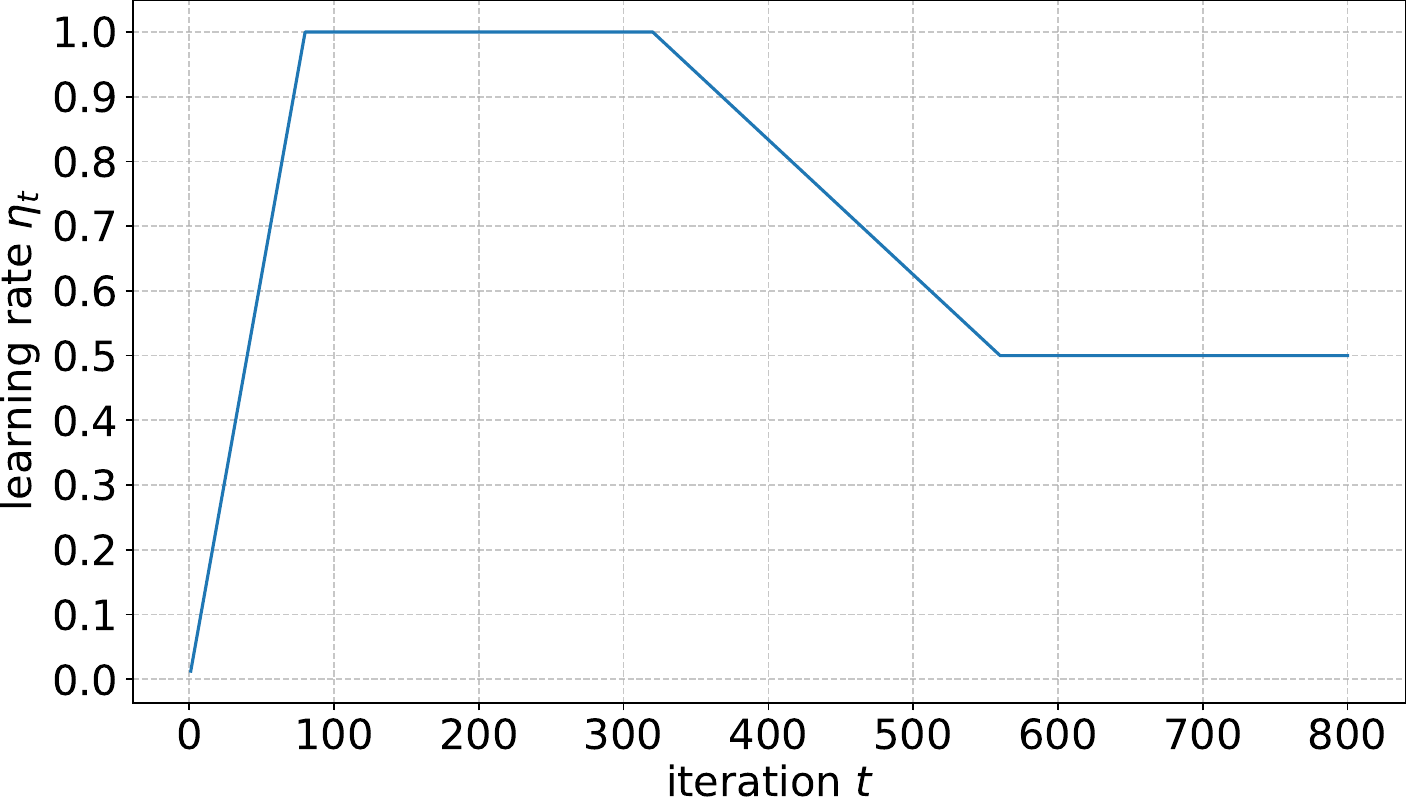}%
        \caption{Function curve between the WSDS learning rate and the number of iterations}
        \label{fig-lr-strategy-wsds}}
    \footnotesize{Note: The horizontal and vertical axes in the figure are set only to intuitively show the trend of the learning rate over the iteration process; the numerical values do not reflect the real data used in the experiments.}
\end{figure}

Learning-rate strategy configuration. Two learning-rate strategies were used for different models: Warmup-Steady-Decay-Steady (WSDS) and Warmup-Steady-Decay (WSD), as shown in Equations~\eqref{eq-lr-strategy-wsds} and~\eqref{eq-lr-strategy-wsd}, respectively:
\begin{equation}
    \eta_t = \begin{cases}
        \eta_{\mathrm{init}} + \frac{\eta_{\mathrm{peak}} - \eta_{\mathrm{init}}}{t_1} t, &\text{if } 0 < t \le t_1, \\
        \eta_{\mathrm{peak}}, &\text{if } t_1 < t \le t_2, \\
        \eta_{\mathrm{peak}} + \frac{\eta_{\min} - \eta_{\mathrm{peak}}}{t_3-t_2} (t-t_2), &\text{if } t_2 < t \le t_3, \\
        \eta_{\min}, &\text{if } t > t_3
    \end{cases}
    \label{eq-lr-strategy-wsds}
\end{equation}
\begin{equation}
    \eta_t = \begin{cases}
        \eta_{\mathrm{init}} + \frac{\eta_{\mathrm{peak}} - \eta_{\mathrm{init}}}{t_1} t, &\text{if } 0 < t \le t_1, \\
        \eta_{\mathrm{peak}}, &\text{if } t_1 < t \le t_2, \\
        \eta_{\mathrm{peak}} + \frac{\eta_{\min} - \eta_{\mathrm{peak}}}{t_3-t_2} (t-t_2), &\text{if } t_2 < t \le t_3
    \end{cases}
    \label{eq-lr-strategy-wsd}
\end{equation}
Here, $\eta_{\mathrm{init}}$ denotes the initial learning rate, $\eta_{\mathrm{peak}}$ denotes the peak learning rate, and $\eta_{\min}$ denotes the minimum learning rate. Figure~\ref{fig-lr-strategy-wsds} visualizes the functional relationship between the WSDS learning rate and the number of iterations.

One difficulty in research on optimization algorithms is how to design experiments that compare different optimizers relatively fairly, because different optimizers have different hyperparameter configurations. Aligning learning rates and weight-decay rates through the RMS norm of the update is relatively fair, namely
\begin{equation}
\begin{cases}
    \eta_{\mathrm{opt_2}} = \eta_{\mathrm{opt_1}} \frac{\lVert \bm{u}_{\mathrm{opt_1}} \rVert_2}{\lVert \bm{u}_{\mathrm{opt_2}} \rVert_2},
    \\
    \lambda_{\mathrm{opt_2}} = \lambda_{\mathrm{opt_1}} \frac{\lVert \bm{u}_{\mathrm{opt_2}} \rVert_2}{\lVert \bm{u}_{\mathrm{opt_1}} \rVert_2}
\end{cases}
    \label{eq-opt-alignment}
\end{equation}
Table~\ref{tbl-opt-lr} lists the peak learning rates of different optimization algorithms on different models. In addition, the initial warmup learning rate of each optimizer was uniformly set to $1 \times 10^{-7}$, and the minimum learning rate was uniformly set to $0.01$ times the peak learning rate. The weight-decay rates of AdamW, RAdam, and Ada2MS were uniformly set to $0.01$, whereas the weight-decay rates of momentum SGD, AdaI, Lion, and SophiaG were computed according to Equation~\eqref{eq-opt-alignment}.

\begin{table}[H]
    \centering
    \caption{Peak learning rates of different optimization algorithms on different models}
    \label{tbl-opt-lr}
    \begin{tblr}{
            width=\textwidth,
            colspec={X[l] X[c] X[c] X[c]},
            hline{1,Z}={1pt},
            hline{2}={0.5pt},
            columns={valign=m},
            rowsep=0.5pt,
            cells={font=\footnotesize}
        }
        & SwinV2-S~\cite{2022-Liu-SwinV2}
        & YOLOv7-tiny~\cite{2023-Wang-YOLOv7}
        & U-Net~\cite{2015-Ronneberger-U-Net} \\
        \SetCell[c=1]{l}{Momentum \\ SGD~\cite{2013-Sutskever-SGDM}} & $2.67 \times 10^{-2}$ & $0.349$ & $0.116$ \\
        AdamW~\cite{2017-Loshchilov-AdamW} & $5.34 \times 10^{-4}$ & $3.49 \times 10^{-3}$ & $1.16 \times 10^{-3}$ \\
        RAdam~\cite{2020-Liu-RAdam} & $5.34 \times 10^{-4}$ & $3.49 \times 10^{-3}$ & $1.16 \times 10^{-3}$ \\
        AdaI~\cite{2022-Xie-AdaptiveInertia} & $1$ & $1$ & $1$ \\
        Lion~\cite{2023-Chen-Lion} & $1.068 \times 10^{-4}$ & $6.98 \times 10^{-4}$ & $2.32 \times 10^{-4}$ \\
        SophiaG~\cite{2024-Liu-Sophia} & $1.068 \times 10^{-4}$ & $6.98 \times 10^{-4}$ & $2.32 \times 10^{-4}$ \\
        \SetCell[c=1]{l}{Ada2MS \\ (Ours)} & $5.34 \times 10^{-4}$ & $3.49 \times 10^{-3}$ & $1.16 \times 10^{-3}$ \\
    \end{tblr}
\end{table}

\subsection{Image classification results and analysis}\label{sec4.2}

Image classification benchmark dataset. CIFAR-100~\cite{2009-Krizhevsky-CIFAR-100} was created by the Department of Computer Science at the University of Toronto in 2009 as an expanded version of the CIFAR series. It contains $60,000$ RGB color images of size $32\times 32$, forming a fine-grained classification framework with $100$ classes grouped into $20$ superclasses, each containing $5$ subclasses. The dataset is strictly split into $50,000$ training images and $10,000$ test images. Random color-jitter transformation and random horizontal flipping were used for data augmentation during training.

Image classification benchmark model. SwinV2~\cite{2022-Liu-SwinV2} is an enhanced visual Transformer developed by Microsoft, with a series of innovations for training large models. To adapt it to the CIFAR resolution, the patch size of SwinV2 was set to $1$ and the window size was set to $4$. The model was reproduced from the official torchvision implementation. Weight initialization followed the official default strategy, and no pretrained parameters were loaded.

Image classification evaluation metrics. Top-1 error rate and Top-5 error rate are standard metrics for evaluating image classification models. The test set contains $N$ images. Let the ground-truth label of the $i$-th image be $y_i$, and let the class-probability vector predicted by the model for that image be $p_i = (p_{i,1}, p_{i,2}, \dots, p_{i,C})$, where $C$ is the total number of classes. The two error rates are defined as follows:
\begin{align}
    &E_{\mathrm{Top}\mbox{-}1} = \frac{1}{N}\sum_{i=1}^{N}
    \chi_A\left(y_i \neq \arg\max_{j} p_{i,j}\right),
    \\
    &E_{\mathrm{Top}\mbox{-}5} = \frac{1}{N}\sum_{i=1}^{N}\chi_A
    \left(y_i \notin \mathrm{top}\mbox{-}5\left({p_{i,j}}_{j=1}^{C}\right)\right)
\end{align}
Here, $\chi_A(.)$ denotes the indicator function. This definition follows the convention of~\cite{2012-Krizhevsky-ImageNetClassification} and has become an evaluation standard for subsequent research.

Optimal performance comparison. Table~\ref{tbl-cifar100-SwinV2-S} reports the results of training SwinV2-S on CIFAR-100 with different optimization algorithms. The number of epochs was set to $160$, the batch size was set to $64$, the WSDS learning-rate strategy was used, and the peak learning rates followed Table~\ref{tbl-opt-lr}. Ada2MS achieved the second-best Top-1 error rate of $29.17\%$, only $0.5\%$ higher than the lowest Top-1 error rate of $28.67\%$.

\begin{table}[H]
    \centering
    \caption{Image classification results of different optimization algorithms on CIFAR-100}
    \label{tbl-cifar100-SwinV2-S}
\begin{threeparttable}
    \begin{tblr}{
            width=\textwidth,
            colspec={X[l] X[c] X[c]},
            hline{1,Z}={1pt},
            hline{2}={0.5pt},
            columns={valign=m},
            rowsep=0.5pt,
            cells={font=\footnotesize}
        }
        & Top-1 error rate ($\%$)$\downarrow$
        & Top-5 error rate ($\%$)$\downarrow$ \\
        Momentum SGD~\cite{2013-Sutskever-SGDM} & 33.22 & 10.73 \\
        AdamW~\cite{2017-Loshchilov-AdamW} & 29.38 & 8.39 \\
        RAdam~\cite{2020-Liu-RAdam} & 29.38 & \underline{8.37} \\
        AdaI~\cite{2022-Xie-AdaptiveInertia} & 31.34 & 10.03 \\
        Lion~\cite{2023-Chen-Lion} & \textbf{28.67} & \textbf{8.17} \\
        SophiaG~\cite{2024-Liu-Sophia} & 29.84 & 8.87 \\
        Ada2MS (Ours) & \underline{29.17} & 8.42 \\
    \end{tblr}
    \begin{tablenotes}
        \scriptsize
        \item Note: Bold denotes the best result, and underline denotes the second-best result.
    \end{tablenotes}
\end{threeparttable}
\end{table}

\subsection{Object detection results and analysis}\label{sec4.3}

Object detection benchmark dataset. Following the standard protocol in PASCAL VOC~\cite{2011-Everingham-VOC2012} object detection, the training data were formed by combining the trainval subsets of VOC2007 and VOC2012, with $16,551$ images covering $20$ object classes. Model evaluation was conducted on the publicly annotated VOC2007 test set, which contains $4,952$ images. Random padding and scaling, color-jitter transformation, and horizontal flipping were used as data augmentation during training.

Object detection benchmark model. YOLOv7-tiny is a lightweight and fast variant of the YOLOv7~\cite{2023-Wang-YOLOv7} series. Based on the compact ELAN-Tiny backbone, it achieves millisecond-level inference while maintaining competitive detection accuracy, balancing accuracy and efficiency. The input image resolution was set to $320\times 320$. Model weight initialization followed the RegNet~\cite{2020-Radosavovic-RegNet} initialization strategy.

Object detection evaluation metric. Mean Average Precision (mAP) is the most widely used evaluation metric in object detection. Its computation is as follows. For each class $j$, all predicted boxes are sorted in descending order by confidence and compared one by one with the ground-truth boxes. If $\mathrm{IoU} \ge T$, for example $T=0.5$ or $0.75$, and the class is correct, the prediction is counted as a True Positive (TP); otherwise, it is counted as a False Positive (FP). Precision and recall are then accumulated as
\begin{align}
    &P_j(k) = \frac{\sum_{i=1}^{k} \mathrm{TP}_i}{k},
    \label{eq-precision}
    \\
    &R_j(k) = \frac{\sum_{i=1}^{k} \mathrm{TP}_i}{N_k}
    \label{eq-recall}
\end{align}
Here, $N_j$ denotes the total number of ground-truth boxes in class $j$, and $k$ denotes the top $k$ predicted boxes. The curve $P_j(R_j)$ is plotted, and the area under this curve gives the average precision (AP) of the class:
\begin{equation}
    \mathrm{AP}_j = \int_{0}^{1} P_j(R_j) \mathrm{d} R_j
    \label{eq-ap}
\end{equation}
Averaging over all classes gives $\mathrm{mAP} = \frac{1}{C} \sum_{j=1}^{C} \mathrm{AP}_j$.

Optimal performance comparison. Table~\ref{tbl-VOC2007-yolov7tiny} reports the results of training YOLOv7-tiny on VOC2007 and VOC2012 with different optimization algorithms. The number of epochs was set to $80$, the batch size was set to $32$, the WSD learning-rate strategy was used, and the peak learning rates followed Table~\ref{tbl-opt-lr}. Ada2MS achieved the highest mAP@0.5 of $59.40\%$ and the highest mAP@0.75 of $35.46\%$.

\begin{table}[H]
    \centering
    \caption{Object detection results of different optimization algorithms on VOC2007 test}
    \label{tbl-VOC2007-yolov7tiny}
\begin{threeparttable}
    \begin{tblr}{
            width=\textwidth,
            colspec={X[l] X[c] X[c]},
            hline{1,Z}={1pt},
            hline{2}={0.5pt},
            columns={valign=m},
            rowsep=0.5pt,
            cells={font=\footnotesize}
        }
        & mAP@0.5 ($\%$)$\uparrow$
        & mAP@0.75 ($\%$)$\uparrow$ \\
        Momentum SGD~\cite{2013-Sutskever-SGDM} & 57.19 & 31.02 \\
        AdamW~\cite{2017-Loshchilov-AdamW} & 59.12 & 34.91 \\
        RAdam~\cite{2020-Liu-RAdam} & 58.89 & 34.66 \\
        AdaI~\cite{2022-Xie-AdaptiveInertia} & 56.91 & 31.31 \\
        Lion~\cite{2023-Chen-Lion} & 56.97 & 32.14 \\
        SophiaG~\cite{2024-Liu-Sophia} & 54.61 & 29.07 \\
        Ada2MS (Ours) & \textbf{59.40} & \textbf{35.46} \\
    \end{tblr}
    \begin{tablenotes}
        \scriptsize
        \item Note: mAP@0.5 denotes $\mathrm{IoU} \ge 0.5$, and mAP@0.75 denotes $\mathrm{IoU} \ge 0.75$.
    \end{tablenotes}
\end{threeparttable}
\end{table}

\subsection{Semantic segmentation results and analysis}\label{sec4.4}

Semantic segmentation benchmark dataset. The Semantic Boundaries~\cite{2011-Hariharan-SBD} dataset contains $11,355$ images from PASCAL VOC 2011. It is split into $8,498$ training images and $2,857$ validation images, and provides pixel-level semantic segmentation annotations and boundary information for $20$ object classes, covering semantic segmentation and boundary detection tasks. Random scaling, horizontal flipping, and random cropping were used as data augmentation during training.

Semantic segmentation benchmark model. U-Net~\cite{2015-Ronneberger-U-Net} was originally proposed for biomedical image segmentation. Its symmetric encoder-decoder architecture uses skip connections to fuse high-resolution details and low-resolution semantic information, achieving accurate pixel-level prediction with limited labeled data. This design balances localization accuracy and contextual modeling capability, making U-Net a widely used baseline model in semantic segmentation. The number of basic units in U-Net was set to $64$, and the input image resolution was set to $320\times 320$. Model weight initialization followed the RegNet~\cite{2020-Radosavovic-RegNet} initialization strategy. The loss function used a weighted average of cross-entropy loss~\cite{1986-Rumelhart-GD-SGD} and Dice loss~\cite{2016-Milletari-DiceLoss}.

Semantic segmentation evaluation metric. Mean Intersection over Union (mIoU) is a standard metric in semantic segmentation. It is computed by calculating the intersection-over-union between the prediction and the ground truth for each class and then taking the average:
\begin{equation}
    \mathrm{mIoU} = \frac{1}{C} \sum_{i=1}^{C} \frac{\mathrm{TP}_i}{\mathrm{TP}_i + \mathrm{FP}_i + \mathrm{FN}_i}
    \label{eq-mIoU}
\end{equation}
Here, $C$ denotes the total number of classes, and $\mathrm{TP}_i$, $\mathrm{FP}_i$, and $\mathrm{FN}_i$ denote the number of true-positive, false-positive, and false-negative pixels for class $i$, respectively. This metric measures both localization precision and classification accuracy; higher values indicate better segmentation performance.

Optimal performance comparison. Table~\ref{tbl-SBD-unet} reports the results of training U-Net on Semantic Boundaries with different optimization algorithms. The number of epochs was set to $80$, the batch size was set to $32$, the WSD learning-rate strategy was used, and the peak learning rates followed Table~\ref{tbl-opt-lr}. Ada2MS achieved the third-best mIoU of $52.52\%$.

\begin{table}[H]
    \centering
    \caption{Semantic segmentation results of different optimization algorithms on Semantic Boundaries}
    \label{tbl-SBD-unet}
\begin{threeparttable}
    \begin{tblr}{
            width=\textwidth,
            colspec={X[l] X[c]},
            hline{1,Z}={1pt},
            hline{2}={0.5pt},
            columns={valign=m},
            rowsep=0.5pt,
            cells={font=\footnotesize}
        }
        & mIoU ($\%$)$\uparrow$ \\
        Momentum SGD~\cite{2013-Sutskever-SGDM} & 52.40 \\
        AdamW~\cite{2017-Loshchilov-AdamW} & 52.28 \\
        RAdam~\cite{2020-Liu-RAdam} & \underline{52.83} \\
        AdaI~\cite{2022-Xie-AdaptiveInertia} & \textbf{53.50} \\
        Lion~\cite{2023-Chen-Lion} & 50.27 \\
        SophiaG~\cite{2024-Liu-Sophia} & 47.92 \\
        Ada2MS (Ours) & \uwave{52.52} \\
    \end{tblr}
    \begin{tablenotes}
        \scriptsize
        \item Note: Wavy underline denotes the third-best result.
    \end{tablenotes}
\end{threeparttable}
\end{table}

\section{Conclusion and future work}\label{sec5}

Ada2MS realizes a smooth transition from AdamW-like behavior to momentum-SGD-like behavior through continuous interpolation between elementwise second-moment normalization and global second-moment normalization. The goal of this mechanism is to gradually introduce more SGD-oriented update characteristics while maintaining update-scale stability as much as possible. Under the current evaluation protocol, Ada2MS obtained competitive results on the CIFAR-100 classification task and achieved the best results among the compared optimizers on the VOC detection task. These results indicate that the mechanism has potential. However, more hybrid baselines, more complete ablation studies of the switching schedule, and validation on non-visual tasks remain necessary in future work.

\section*{Acknowledgments}

This work was partly supported by the Jiangxi Provincial Key Laboratory of Virtual Reality (Grant No. 2024SSY03151).

\section*{Data availability}

The datasets during the current study are open source and available in \url{http://www.cs.toronto.edu/~kriz/cifar.html},  \url{https://www.kaggle.com/datasets/zaraks/pascal-voc-2007}, \url{https://www.kaggle.com/datasets/huanghanchina/pascal-voc-2012}, 
\url{https://www2.eecs.berkeley.edu/Research/Projects/CS/vision/grouping/semantic_contours/benchmark.tgz} repositories, respectively.


\begin{thebibliography}{35}
	\expandafter\ifx\csname url\endcsname\relax
	\def\url#1{\texttt{#1}}\fi
	\expandafter\ifx\csname urlprefix\endcsname\relax\def\urlprefix{URL }\fi
	\expandafter\ifx\csname href\endcsname\relax
	\def\href#1#2{#2} \def\path#1{#1}\fi
	
	\bibitem{2021-Jumper-AlphaFold2}
	J.~Jumper, R.~Evans, A.~Pritzel, et~al., Highly accurate protein structure
	prediction with alphafold, Nature 596~(7873) (2021) 583--589.
	\newblock \href {http://dx.doi.org/10.1038/s41586-021-03819-2}
	{\path{doi:10.1038/s41586-021-03819-2}}.
	
	\bibitem{2023-Merchant-scaling}
	A.~Merchant, S.~Batzner, S.~S. Schoenholz, et~al., Scaling deep learning for
	materials discovery, Nature 624~(7990) (2023) 80--85.
	\newblock \href {http://dx.doi.org/10.1038/s41586-023-06735-9}
	{\path{doi:10.1038/s41586-023-06735-9}}.
	
	\bibitem{2024-Kim-HFP}
	S.~K. Kim, R.~Shousha, S.~M. Yang, et~al., Highest fusion performance without
	harmful edge energy bursts in tokamak, Nature Communications 15~(1) (2024)
	3990--4001.
	\newblock \href {http://dx.doi.org/10.1038/s41467-024-48415-w}
	{\path{doi:10.1038/s41467-024-48415-w}}.
	
	\bibitem{2023-Touvron-Llama2}
	H.~Touvron, L.~Martin, K.~Stone, et~al.,
	\href{https://arxiv.org/abs/2307.09288}{Llama 2: Open foundation and
		fine-tuned chat models} (2023).
	\newline\urlprefix\url{https://arxiv.org/abs/2307.09288}
	
	\bibitem{2017-Loshchilov-AdamW}
	I.~Loshchilov, F.~Hutter, \href{http://arxiv.org/abs/1711.05101}{Fixing weight
		decay regularization in adam} (2017).
	\newline\urlprefix\url{http://arxiv.org/abs/1711.05101}
	
	\bibitem{2017-Keskar-SwitchingAdamSGD}
	N.~S. Keskar, R.~R. Socher, \href{http://arxiv.org/abs/1712.07628}{Improving
		generalization performance by switching from adam to sgd} (2017).
	\newline\urlprefix\url{http://arxiv.org/abs/1712.07628}
	
	\bibitem{2022-Xie-AdaptiveInertia}
	Z.~Xie, X.~Wang, H.~Zhang, et~al., Adaptive inertia: Disentangling the effects
	of adaptive learning rate and momentum, in: International Conference on
	Machine Learning, Vol. 162, 2022, pp. 24430--24459.
	
	\bibitem{2013-Sutskever-SGDM}
	I.~Sutskever, J.~Martens, G.~Dahl, G.~Hinton, On the importance of
	initialization and momentum in deep learning, in: International Conference on
	Machine Learning, 2013, p. 1139–1147.
	
	\bibitem{2021-Dosovitskiy-ViT}
	A.~Dosovitskiy, L.~Beyer, A.~Kolesnikov, et~al., An image is worth 16x16 words:
	Transformers for image recognition at scale, in: International Conference on
	Learning Representations, 2021, pp. 1--21.
	
	\bibitem{2022-Liu-ConvNeXt}
	Z.~Liu, H.~Mao, C.~Y. Wu, et~al., A convnet for the 2020s, in: IEEE/CVF
	Conference on Computer Vision and Pattern Recognition, 2022, pp.
	11966--11976.
	\newblock \href {http://dx.doi.org/10.1109/CVPR52688.2022.01167}
	{\path{doi:10.1109/CVPR52688.2022.01167}}.
	
	\bibitem{1986-Rumelhart-GD-SGD}
	D.~E. Rumelhart, G.~E. Hinton, R.~J. Williams, Learning representations by
	back-propagating errors, Nature 323~(6088) (1986) 533--536.
	\newblock \href {http://dx.doi.org/10.1038/323533a0}
	{\path{doi:10.1038/323533a0}}.
	
	\bibitem{2019-Gitman-UnderstandingSGDM}
	I.~Gitman, H.~Lang, P.~Zhang, L.~Xiao, Understanding the role of momentum in
	stochastic gradient methods, in: International Conference on Neural
	Information Processing Systems, Vol.~32, 2019, pp. 1--11.
	
	\bibitem{2024-Ramezani-Kebrya-GeneralizationSGDM}
	A.~Ramezani-Kebrya, K.~Antonakopoulos, V.~Cevher, et~al., On the generalization
	of stochastic gradient descent with momentum, Journal of Machine Learning
	Research 25~(22) (2024) 1--56.
	
	\bibitem{2024-Zhao-SNGDM}
	S.~Zhao, C.~Shi, Y.~Xie, W.~Li, Stochastic normalized gradient descent with
	momentum for large-batch training, SCIENCE CHINA Information Sciences 67~(11)
	(2024) 1--15.
	\newblock \href {http://dx.doi.org/10.1007/s11432-022-3892-8}
	{\path{doi:10.1007/s11432-022-3892-8}}.
	
	\bibitem{2011-Duchi-AdaGrad}
	J.~Duchi, E.~Hazan, Y.~Singer, Adaptive subgradient methods for online learning
	and stochastic optimization, The Journal of Machine Learning Research 12
	(2011) 2121--2159.
	
	\bibitem{2012-Tieleman-RMSProp}
	T.~Tieleman, G.~Hinton,
	\href{https://cir.nii.ac.jp/crid/1370017282431050757}{Rmsprop: Divide the
		gradient by a running average of its recent magnitude} (2012).
	\newline\urlprefix\url{https://cir.nii.ac.jp/crid/1370017282431050757}
	
	\bibitem{2015-Kingma-Adam}
	D.~P. Kingma, J.~Ba, Adam: A method for stochastic optimization, in:
	International Conference on Learning Representations, 2015, pp. 1--15.
	
	\bibitem{2018-Reddi-AMSGrad}
	S.~J. Reddi, S.~Kale, S.~Kumar, On the convergence of adam and beyond, in:
	International Conference on Learning Representations, 2018, pp. 1--23.
	
	\bibitem{2018-Shazeer-Adafactor}
	N.~Shazeer, M.~Stern, Adafactor: Adaptive learning rates with sublinear memory
	cost, in: International Conference on Machine Learning, Vol.~80, 2018, pp.
	4596--4604.
	
	\bibitem{2024-Liu-Sophia}
	H.~Liu, Z.~Li, D.~L.~W. Hall, et~al., Sophia: A scalable stochastic
	second-order optimizer for language model pre-training, in: International
	Conference on Learning Representations, 2024, pp. 1--30.
	
	\bibitem{2020-Liu-RAdam}
	L.~Liu, H.~Jiang, P.~He, et~al., On the variance of the adaptive learning rate
	and beyond, in: International Conference on Learning Representations, 2020,
	pp. 1--13.
	
	\bibitem{2020-Li-Adax}
	W.~Li, Z.~Zhang, X.~Wang, P.~Luo,
	\href{https://openreview.net/forum?id=r1l-5pEtDr}{Adax: Adaptive gradient
		descent with exponential long term memory} (2020).
	\newline\urlprefix\url{https://openreview.net/forum?id=r1l-5pEtDr}
	
	\bibitem{2025-Zhu-AdamNX}
	M.~Zhu, Q.~Xiao, W.~Min, \href{https://arxiv.org/abs/2511.13465}{Adamnx: An
		adam improvement algorithm based on a novel exponential decay mechanism for
		the second-order moment estimate} (2025).
	\newline\urlprefix\url{https://arxiv.org/abs/2511.13465}
	
	\bibitem{2023-Chen-Lion}
	X.~Chen, C.~Liang, D.~Huang, et~al., Symbolic discovery of optimization
	algorithms, in: International Conference on Neural Information Processing
	Systems, 2023, pp. 1--30.
	
	\bibitem{2024-Jordan-Muon}
	K.~Jordan, Y.~Jin, V.~Boza, et~al.,
	\href{https://kellerjordan.github.io/posts/muon}{Muon: An optimizer for
		hidden layers in neural networks} (2024).
	\newline\urlprefix\url{https://kellerjordan.github.io/posts/muon}
	
	\bibitem{2009-Krizhevsky-CIFAR-100}
	A.~Krizhevsky, G.~Hinton, \href{https://citeseerx.ist.psu.edu}{Learning
		multiple layers of features from tiny images} (2009).
	\newline\urlprefix\url{https://citeseerx.ist.psu.edu}
	
	\bibitem{2011-Everingham-VOC2012}
	M.~Everingham, J.~Winn, The pascal visual object classes challenge 2012
	(voc2012) development kit, Pattern Analysis, Statistical Modelling and
	Computational Learning, Tech. Rep 8.
	
	\bibitem{2011-Hariharan-SBD}
	B.~Hariharan, P.~Arbeláez, L.~Bourdev, et~al., Semantic contours from inverse
	detectors, in: IEEE/CVF International Conference on Computer Vision, 2011,
	pp. 991--998.
	\newblock \href {http://dx.doi.org/10.1109/ICCV.2011.6126343}
	{\path{doi:10.1109/ICCV.2011.6126343}}.
	
	\bibitem{2022-Liu-SwinV2}
	Z.~Liu, H.~Hu, Y.~Lin, et~al., Swin transformer v2: Scaling up capacity and
	resolution, in: IEEE/CVF Conference on Computer Vision and Pattern
	Recognition, 2022, pp. 12009--12019.
	
	\bibitem{2023-Wang-YOLOv7}
	C.~Wang, A.~Bochkovskiy, H.~Liao, Yolov7: Trainable bag-of-freebies sets new
	state-of-the-art for real-time object detectors, in: IEEE/CVF Conference on
	Computer Vision and Pattern Recognition, 2023, pp. 7464--7475.
	\newblock \href {http://dx.doi.org/10.1109/CVPR52729.2023.00721}
	{\path{doi:10.1109/CVPR52729.2023.00721}}.
	
	\bibitem{2015-Ronneberger-U-Net}
	O.~Ronneberger, P.~Fischer, T.~Brox, U-net: Convolutional networks for
	biomedical image segmentation, in: Medical Image Computing and
	Computer-Assisted Intervention, 2015, pp. 234--241.
	
	\bibitem{2019-Paszke-PyTorch}
	A.~Paszke, S.~Gross, F.~Massa, et~al., Pytorch: An imperative style,
	high-performance deep learning library, in: International Conference on
	Neural Information Processing Systems, Vol.~32, 2019, pp. 1--12.
	
	\bibitem{2012-Krizhevsky-ImageNetClassification}
	A.~Krizhevsky, I.~Sutskever, G.~E. Hinton, Imagenet classification with deep
	convolutional neural networks, in: International Conference on Neural
	Information Processing Systems, Vol.~25, 2012, pp. 1--9.
	
	\bibitem{2020-Radosavovic-RegNet}
	I.~Radosavovic, R.~P. Kosaraju, R.~Girshick, et~al., Designing network design
	spaces, in: IEEE/CVF Conference on Computer Vision and Pattern Recognition,
	2020, pp. 10428--10436.
	
	\bibitem{2016-Milletari-DiceLoss}
	F.~Milletari, N.~Navab, S.~A. Ahmadi, V-net: Fully convolutional neural
	networks for volumetric medical image segmentation, in: IEEE/CVF
	International Conference on 3D Vision, 2016, pp. 565--571.
	\newblock \href {http://dx.doi.org/10.1109/3DV.2016.79}
	{\path{doi:10.1109/3DV.2016.79}}.
	
\end{thebibliography}

\end{document}